\documentclass[final]{l4dc2024}

\usepackage{csvsimple}
\usepackage{multirow}
\usepackage{xcolor}
\usepackage{mathabx}
\usepackage{pifont}
\usepackage{bm, dsfont}

\title[Global Rewards in MADRL for AMoD]{Global Rewards in Multi-Agent Deep Reinforcement Learning for Autonomous Mobility on Demand Systems}
\usepackage{times}
\author{
 \Name{Heiko Hoppe} \Email{heiko.hoppe@tum.de}\\
 \addr Technical University of Munich, Germany
 \AND
 \Name{Tobias Enders} \Email{tobias.enders@tum.de}\\
 \addr Technical University of Munich, Germany
 \AND
 \Name{Quentin Cappart} \Email{quentin.cappart@polymtl.ca}\\
 \addr Polytechnique Montréal, Canada
 \AND
 \Name{Maximilian Schiffer} \Email{schiffer@tum.de}\\
 \addr Technical University of Munich, Germany
}

\begin{document}

\maketitle

\begin{abstract}
We study vehicle dispatching in autonomous mobility on demand (AMoD) systems, where a central operator assigns vehicles to customer requests or rejects these with the aim of maximizing its total profit. Recent approaches use multi-agent deep reinforcement learning (MADRL) to realize scalable yet performant algorithms, but train agents based on local rewards, which distorts the reward signal with respect to the system-wide profit, leading to lower performance. We therefore propose a novel global-rewards-based MADRL algorithm for vehicle dispatching in AMoD systems, which resolves so far existing goal conflicts between the trained agents and the operator by assigning rewards to agents leveraging a counterfactual baseline. Our algorithm shows statistically significant improvements across various settings on real-world data compared to state-of-the-art MADRL algorithms with local rewards. We further provide a structural analysis which shows that the utilization of global rewards can improve implicit vehicle balancing and demand forecasting abilities. Our code is available at \url{https://github.com/tumBAIS/GR-MADRL-AMoD}.
\end{abstract}


\section{Introduction}
Within the coming years, autonomous mobility on demand (AMoD) promises to significantly improve urban passenger transportation. An AMoD system enables its operator to exercise full control over the vehicle fleet, thus improving vehicle dispatching performance, provided that the operator uses an effective control algorithm. This ultimately yields increased vehicle utilization while preserving the comfort of individual mobility. By deciding which requests to accept and assigning vehicles to these requests, the operator faces a contextual multi-stage stochastic control problem. Various approaches to solve this problem exist, ranging from model predictive control \citep[e.g.,][]{AlonsoMora2017} to deep reinforcement learning (DRL) \citep[e.g.,][]{Xu2018, Wang2018, Enders2023}. The latter shows better performance due to its model-free nature. In this context, multi-agent approaches allow to achieve scalability to large instances. To train the agents' policy, existing work uses local, egoistic, per-agent rewards, which can lead to sub-optimal behavior from the perspective of the central operator interested in maximizing the system-wide profit.

In this work, we improve upon the state-of-the-art local-rewards algorithm (LRA) of \citet{Enders2023} by proposing a novel way to train agents with global rewards. The core of our methodology is a new advantage function, which estimates the individual agents' contribution to the global reward, as such aligning their goal with the central operator, ultimately following a system-optimal policy. Our algorithm outperforms LRA by up to 2\% on average across test dates and up to 6\% on individual dates, which is a substantial performance improvement in AMoD. Additionally, our algorithm is as scalable as LRA, accordingly setting a new state-of-the-art for vehicle dispatching in AMoD systems.

\subsection{Related work}
Our work relates to two research streams: from an application perspective to AMoD fleet control, and methodologically to multi-agent deep reinforcement learning (MADRL). We review related literature from both streams in the following.

Algorithms for (A)MoD fleet control range from greedy heuristics \citep[e.g.,][]{Liao2003, Lee2004}, to (stochastic) model predictive control \citep[e.g.,][]{AlonsoMora2017, Tsao2018}, combinations of optimization and supervised learning \citep[e.g.,][]{Zhang2017, Jungel2023}, and DRL. Existing DRL-based algorithms differ from our approach, as they often use value-based learning \citep[e.g.,][]{Xu2018, Wang2018, SadeghiEshkevari2022, MenesesCime2022}, which handles large discrete action spaces less efficiently \citep[cf. ][]{Akkerman2024}. Others focus on explicit rebalancing instead of request assignment \citep[e.g.,][]{Jiao2021, Liang2022, He2022}. Furthermore, most previous work considers non-autonomous MoD, aiming to maximize revenues \citep[e.g.,][]{Wang2018, Xu2018, Tang2019}, while operators of AMoD systems commonly focus on maximizing operational profit \citep[cf.][]{Enders2023}. Finally, the  work of \citet{Enders2023} uses policy-based hybrid MADRL for a profit-maximizing AMoD operator, but trains agents on local rewards, possibly leading to sub-optimal agent behavior. To the best of our knowledge, no work exists that addresses the problem of profit-maximizing vehicle dispatching in AMoD without the aforementioned shortcomings. We address this research gap by extending the work of \citet{Enders2023} to incorporate global rewards, thus creating goal congruence between the agents and the system operator.

A crucial challenge in our setting is deriving per-agent contributions to global success from a shared reward signal, i.e., a credit assignment problem \citep{Weiss1995, Wolpert1999, Chang2003}. Solution approaches like inverse reinforcement learning \citep[e.g.,][]{Ng2000, HadfieldMenell2017, Lin2018} or value decomposition \citep[e.g.,][]{Kok2006, Sunehag2018, Son2019, Rashid2020} are not applicable to our setting, because we cannot observe the behavior of an optimal agent and do not use Q-learning. An alternative approach is reward marginalization, based on difference rewards \citep[][]{WolpertTumer2001}, which uses functions (e.g., advantage functions) to estimate the contribution of individual agents to global rewards \citep[e.g.,][]{Nguyen2018, Wu2018, Foerster2018}. Reward marginalization approaches often use actor-critic algorithms. However, reward marginalization approaches often suffer from high learning variance and poor sample efficiency, as they are typically built on top of basic DRL algorithms. To the best of our knowledge, no work exists that combines reward marginalization approaches with low-variance actor-critic algorithms so far. We address this research gap by embedding reward marginalization into a Soft Actor-Critic (SAC) framework for discrete actions.

\subsection{Contribution}
To close the research gap outlined above, we develop the first scalable MADRL-based algorithm for vehicle dispatching in profit-maximizing AMoD that trains agents via global rewards. In this context, we combine SAC for discrete actions with credit assignment based on a counterfactual baseline to resolve goal conflicts between the trained agents and the operator's global objective. Our algorithm combines the benefits of low learning variance and sample efficiency of SAC with the benefits of credit assignment via a counterfactual baseline. Since this algorithm based on purely global rewards scales only to medium-sized problem instances, we additionally develop a scheduled algorithm that combines local rewards and global rewards with our counterfactual baseline. Thus, we obtain a powerful new MADRL algorithm with possible applications beyond AMoD. We evaluate our algorithm on real-world taxi data and show that it outperforms LRA of \citet{Enders2023} by up to 2\% on average across test dates and up to 6\% on individual dates. This constitutes a substantial performance improvement in an AMoD context, where a single percent improvement yields significant daily monetary gains. We further provide a structural analysis which shows that the utilization of global rewards can improve implicit vehicle balancing and demand forecasting abilities.

\section{Problem setting}

\begin{figure}[]
    \vspace{-8 mm}
    \includegraphics[width=\textwidth, keepaspectratio]{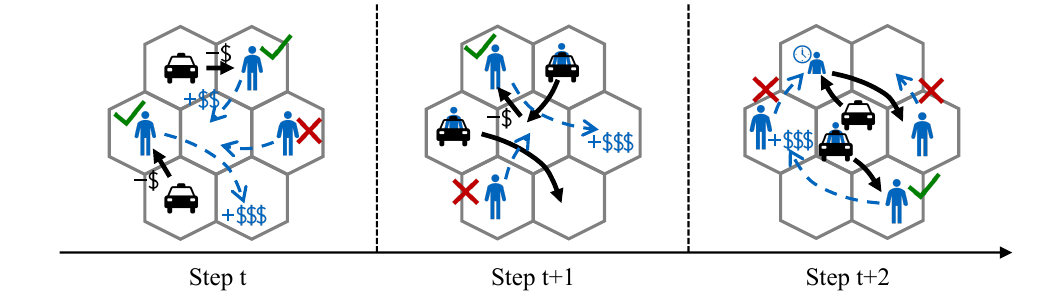}
    \vspace{-10 mm}
    \caption{Exemplary vehicle dispatching process.}
    \label{fig:decision}
    \vspace{-6 mm}
\end{figure}

We focus on the contextual multi-stage stochastic control problem from \citet{Enders2023}, illustrated in Figure~\ref{fig:decision}, to ensure comparability. In this problem setting, a central operator controls an AMoD fleet and dispatches vehicles to serve requests. We consider a discrete time horizon. During each time step, customers submit a variable number of new requests for point-to-point transportation, of which the operator has no prior knowledge. At the beginning of each time step, the operator makes a decision for the batch of requests that were placed during the previous time step. For each request, this decision is either to reject the request or to assign it to a vehicle. The operator can base its decision on fully observable state information, including the requests' origins and destinations as well as the vehicles' positions and already assigned but not yet completed requests. Each request has to be decided upon immediately and customers have a fixed maximum waiting time until they must be picked up. Based on the operator's decision, the system transitions to the next state: vehicles move towards their next destination and potentially pick up or drop off customers. Customers place new requests, while rejected requests leave the system. We simulate the requests by replaying historical data. Finally, the operator receives the system-wide profit as a reward. The profit is calculated as the sum of all revenues from accepted requests minus the costs for all vehicle movements. Revenues and costs are linear functions of the distance travelled. Since fixed costs do not depend on the dispatching problem, we only include operational costs (e.g., for fuel and maintenance). For further details, including a formal Markov decision process, we refer to Appendix \ref{app:problem}.

\section{Methodology}
In the following, we first motivate our baseline algorithm \citep[cf.][]{Enders2023} and extend it with a naive global reward allocation scheme. We then focus on reward marginalization and show how to modify the existing Counterfactual Multi-Agent Policy Gradient (COMA) paradigm to SAC architectures. We finally discuss enhancements to scale our algorithm to large-scale instances.

\begin{figure}[]
    \vspace{-6 mm}
    \includegraphics[width=\textwidth, keepaspectratio]{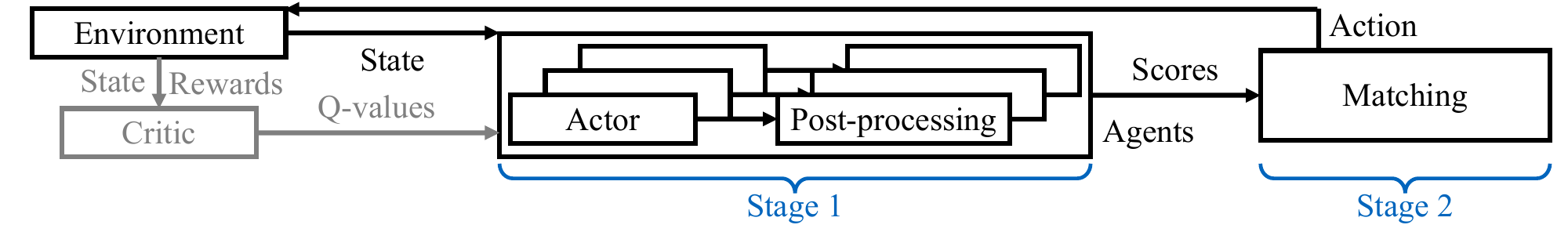}
    \vspace{-10 mm}
    \caption{Outline of base algorithm. Black parts are used during training and testing, gray parts only during training.}
    \label{fig:algorithm}
    \vspace{-10 mm}
\end{figure}

\subsection{Base algorithm}\label{sec:base_algo}

Our problem setting has two peculiarities influencing the construction of an effective vehicle dispatching algorithm: the global action space is too large to use single-agent DRL and the number of requests varies between time steps. When addressing the first peculiarity with multi-agent learning, the need for agent coordination arises. The algorithm of \citet{Enders2023} therefore uses a two-stage architecture, see Figure~\ref{fig:algorithm}. The first stage consists of DRL-trained agents, the second one of an optimization-based central matching. In the first stage, each agent is the combination of a vehicle and a request. Therefore, there exists an agent for every possible vehicle-request combination. An agent can either accept the assignment of the respective request to the respective vehicle or reject this assignment, resulting in a per-agent action space of size two. As the number of requests varies, so does the number of agents. Each agent observes the global state, using an encoding and an attention mechanism to cope with the varying number of requests. Using an actor neural network with a softmax function, the agent obtains probabilities for acceptance and rejection. The per-agent post-processing selects the action by sampling during training or taking the action with the highest probability during testing. It then sets the agent's score to the acceptance probability in case of an acceptance and to zero in case of a rejection. The scores of all agents are submitted to the global matching, which coordinates the agents. This bipartite matching selects the global action by maximizing the submitted scores under the constraints that every request can be assigned at most once and that every vehicle can take at most one new request.

We train the agents using SAC for discrete actions \citep{Haarnoja2018, Christodoulou2019}. As the use of multiple agents only serves to handle the large action space, we train a single actor and a single critic network and use them for all agents. The rewards we use for training are the profits obtained by the algorithm: when a request is assigned to a vehicle and served within the maximum waiting time, the respective agent immediately receives the profit as a reward.

To transition from this algorithm to  our contribution, we first follow a naive approach to include global rewards in training by replacing the per-agent rewards with global rewards when training the critic, resulting in a basic global-rewards algorithm (GRA). We obtain global rewards by summing the profits of all agents at one time step. As this number can be substantially larger than per-agent rewards, we divide it by the average number of non-zero rewards per observation in the replay buffer to stabilize learning. Straightforwardly using this approach leads to a credit assignment problem, as the reward given to agents now depends on the actions of all agents, which generally complicates an agent's learning task. To mitigate this, we explore a credit assignment paradigm in the following.

\subsection{Naive COMA}\label{sec:naive_COMA}

A suitable credit assignment paradigm for our setting is COMA, proposed by \citet{Foerster2018}. COMA fits our setting best, because it has a similar structure as SAC and is especially suitable for small per-agent action spaces. Following the main rationale of COMA, agents should maximize their contribution to the global reward instead of maximizing the global reward directly. As obtaining global rewards for several actions is computationally infeasible, COMA trains the critic on global rewards to approximate global state-action-values. The contribution of an agent to this global value is defined as the value of taking an action in contrast to the value of taking a default action. This default action is calculated as the policy-weighted average value of all possible actions the agent can take (counterfactual baseline). The advantage function of COMA therefore is

\begin{equation}
    A_i(a_i|s,i)=Q_\theta(a_i|s,\bar a_{-i})-\sum_{a'_i}\pi_\phi(a'_i|s,i) \:Q_\theta(a'_i|s,\bar a_{-i}).
\end{equation}
In this and all following equations, $s$ denotes the global state, $a'_i$ an action agent~$i$ can take as a reject/accept decision before the global matching, $a_i$ the action the agent actually takes, and $\bar a_{-i}$ the actions of all agents except agent~$i$ after the global matching. Let $\pi_\phi(a_i|s,i) \in [0,1]$ be the probability that agent~$i$ takes action~$a_i$ in state~$s$, following policy~$\pi_\phi$ parameterized by network $\phi$. $Q_\theta(a_i|s,\bar a_{-i}) \in \mathbb{R}^2$ is the global state-action-value (Q-value) of action~$a_i$ taken by agent~$i$ in state~$s$ estimated using network~$\theta$, given other agents' actions~$\bar a_{-i}$.

\citet{Foerster2018} use a sampling-based approach to estimate the actor's loss function and base its computation only on the action taken by the agent. The loss function thus reads $J_\pi(\phi)=\mathbb{E}_{s\sim D}\bigl[ \sum_i A_i(a_i|s,i)\bigr]$, with~$D$ denoting the replay buffer. In contrast to that, \citet{Enders2023} use SAC with discrete actions, which considers all possible actions of an agent in the loss function
\begin{equation}\label{eqn:ego_loss}
    J_\pi(\phi)=\mathbb{E}_{s\sim D}\left[ \sum_i \sum_{a_i} \pi_\phi(a_i|s,i) \left(\alpha\log\pi_\phi(a_i|s,i)-\min_{j \in \{1,2\}} \left\{ Q_\theta^j(a_i|s,\bar a_{-i}) \right\} \right) \right],
\end{equation}
with $\alpha\in\mathbb{R}$ being the entropy coefficient. To use credit assignment via COMA in combination with the proven reliability and low variance of SAC, we need to integrate the baseline of COMA into the loss function of SAC. In the following, we use $\pi(a_i):=\pi_\phi(a_i|s,i)$ and $Q(a_i):=\min_{j \in \{1,2\}} \left\{ Q_\theta^j(a_i|s,\bar a_{-i}) \right\}$ for conciseness. Then, considering one instance of a batch and one agent, the loss function of SAC extended by the advantage of COMA reads
\begin{equation}\label{eqn:loss}
    J_\pi(\phi|s,i)=\sum_{a_i}\pi(a_i) \left(\alpha\log\pi(a_i)-Q(a_i)+\sum_{a'_i}\pi(a'_i)\: Q(a'_i) \right).
\end{equation}
\begin{proposition}
    The loss function $J_\pi(\phi|s,i)$ as defined in Equation~\eqref{eqn:loss} is equivalent to the entropy $J_\pi(\phi|s,i)=\sum_{a_i}\pi(a_i)\: \alpha\log\pi(a_i)$ of a plain SAC architecture.
    \label{prop:equivalence}
\end{proposition}
For a proof of Proposition~\ref{prop:equivalence}, we refer to Appendix \ref{sec:proof}.
Accordingly, using the loss function as derived in Equation~\eqref{eqn:loss} does not allow to learn a meaningful policy such that we cannot simply apply the COMA paradigm to our SAC framework. This observation motivates us to study a novel approach to combine SAC for discrete actions with COMA to learn a good policy with global rewards.

\subsection{Adjusted COMA for SAC architectures}\label{sec:approach}

To solve the convergence problem outlined above, we have to adjust the loss function in Equation \eqref{eqn:loss}. Using only the action taken by the agent for the loss function as in \citet{Foerster2018} does not lead to convergence even in small experimental instances, as it increases the loss function's variance. We therefore adjust $\pi(a'_i)$, changing the weighting of the default action in the baseline. This is possible from a theoretical perspective, as the exact specification of a default action is not derived from the idea of difference rewards \citep{WolpertTumer2001}, but left to the user's discretion.

Straightforwardly, we can define the default action by using an equally-weighted average instead of a policy-weighted average, resulting in the advantage function $A^\text{equ}_i(a_i)=Q(a_i)-\sum_{a'_i}\frac{1}{n_{a'_i}}\: Q(a'_i)$, with~$n_{a_i}$ being the number of actions per agent. We call this algorithm $\text{COMA}^\text{equ}$, which resolves the convergence problem, but has a disadvantage: when the actor network estimates  different probabilities for single actions, weighting all actions equally is not a reasonable default action. This problem is especially pronounced during late training, when the actor network is better at estimating action probabilities. We solve this issue by defining a second default action with the use of a target actor network~$\bar\phi$. This network has the same structure and initialization as the actor network~$\phi$ and is updated using exponential averages of the actor network's parameters, similar to target networks in Q-learning. The advantage function of this algorithm ($\text{COMA}^\text{tgt}$) is $A^\text{tgt}_i(a_i)=Q(a_i)-\sum_{a'_i}\pi_{\bar\phi}(a'_i)\: Q(a'_i)$. Since $\bar\phi$ differs from $\phi$, this algorithm solves the convergence problem as well. During early training, $\text{COMA}^\text{tgt}$ is not as suitable as $\text{COMA}^\text{equ}$, since the sub-optimal action probabilities of an untrained target actor network are a disadvantage compared to equally-weighted actions. Later in training, $\bar\phi$ can estimate better action probabilities, making $\text{COMA}^\text{tgt}$ superior to $\text{COMA}^\text{equ}$.

Since we now have one algorithm especially suitable for early learning and one especially suitable for later learning, we combine these two and obtain $\text{COMA}^\text{adj}$, based on a dynamic combination of the two newly introduced advantage functions. The advantage function of $\text{COMA}^\text{adj}$ thus reads
\begin{equation} \label{eqn:COMA_adj}
\begin{split}
    A^\text{adj}_i(a_i)&=(1-\beta)A^\text{equ}_i(a_i)+\beta A^\text{tgt}_i(a_i) \\
    &=Q(a_i)-(1-\beta)\sum_{a'_i}\frac{1}{n_{a_i}}\: Q(a'_i)-\beta\sum_{a'_i}\pi_{\bar\phi}(a'_i)\: Q(a'_i).
\end{split}
\end{equation}
Here, the hyperparameter $\beta\in[0,1]$ is the weight of the $\text{COMA}^\text{tgt}$ baseline and follows a schedule depending on the training iteration, starting at zero and ending at one. Using a simple linear schedule usually works best, see Appendix~\ref{app:schedules}. Then, the loss function of $\text{COMA}^\text{adj}$ reads
\begin{equation}
    J^\text{adj}_\pi(\phi|s,i)=\sum_{a_i}\pi(a_i) \left(\alpha\log\pi(a_i)-A^\text{adj}_i(a_i)\right).
\end{equation}

With this loss function, $\text{COMA}^\text{adj}$ solves the credit assignment problem. In our experiments, $\text{COMA}^\text{adj}$ performs better than LRA, but has a scalability problem: when the number of agents increases beyond medium-sized problem instances, $\text{COMA}^\text{adj}$ fails to converge. Reasons for this are the diminishing influence of a single agent on global rewards and the overlap of many agents' actions when the number of agents increases, making learning per-agent Q-values difficult \citep[cf.][]{Rashid2020}. We therefore investigate how to scale $\text{COMA}^\text{adj}$.

\subsection{Reward scheduling}\label{sec:scalability}

Usually, one could resolve the scalability problem of $\text{COMA}^\text{adj}$ straightforwardly by adjusting the critic to accommodate value factorization \citep[e.g.,][]{Su2021}, but this approach is infeasible in our setting as the number of agents is variable. Similarly, learning the critic on a static mix of local and global rewards in a local-global-rewards algorithm (LGRA) does not solve the scalability problem, since any non-negligible share of global rewards distorts learning when increasing the number of agents. In addition, reward marginalization with a counterfactual baseline is problematic for partially local rewards.

Instead, we can train a single actor network using a weighted average policy loss function, consisting of the loss function for LRA, $J^\text{loc}_\pi (\phi|s,i)$, by \citet{Enders2023} and $\text{COMA}^\text{adj}$. The loss function thus reads
\begin{equation}
    J^\text{scd}_\pi(\phi|s,i)=(1-\kappa)\:J^\text{loc}_\pi (\phi|s,i)+\kappa\:J^\text{adj}_\pi(\phi|s,i),
\end{equation}
with $\kappa\in[0,1]$ being the weight of the loss function of $\text{COMA}^\text{adj}$. Again, $\kappa$ follows a schedule depending on the training iteration, increasing linearly, following a power function or jumping from zero to one at a specified point. Power functions with exponents between 0.01 and 0.5 generally work best, with larger exponents being more suitable for large instance sizes, see Appendix~\ref{app:schedules}. This leads to a new algorithm, we call it $\text{COMA}^\text{scd}$. $\text{COMA}^\text{scd}$ solves the scalability problem, as it enables the learning of global Q-values when increasing the number of agents. The reason this works is the utilization of experience collected following a mixed policy: this way, more diverse experience is available than if using solely own experience, thus improving learning without the destabilizing influence of increasing the entropy. We can therefore train four critic networks from the beginning on, two for local and two for global rewards. Two networks each are necessary for SAC, where we always use the minimum of the two state-action-values to avoid value overestimation. Due to the influence of both local and global rewards, $\text{COMA}^\text{scd}$ can sometimes have a lower performance than algorithms purely based on global rewards, but makes up for this by being as scalable as LRA, thus sacrificing a portion of its performance for scalability.

\section{Numerical studies}

We benchmark our algorithms using the experimental design of \citet{Enders2023}, which bases on New York Taxi data \citep{NYCTLC2015} in a hexagonal grid of Manhattan, with 38 large, 11 small, or 5 small zones. In this realm, we study five instances: two edge cases with high (5 zones, 15 vehicles) and low (11 zones, 6 vehicles) acceptance rates, two typical test cases (11 zones, 18 and 24 vehicles), and a comparatively large instance (38 zones, 100 vehicles). For details on the experimental design, we refer to Appendix \ref{app:setup} and report hyperparameters for all models in Appendix \ref{app:hyperparameters}.

Firstly, we test $\text{COMA}^\text{scd}$ on all five instances and benchmark it against LRA of \citet{Enders2023} and a greedy algorithm, which considers requests in their order of submission, accepting profitable ones and rejecting all others. Secondly, we present an ablation study to show the superiority of $\text{COMA}^\text{scd}$ over our alternative algorithms that use global rewards. Thirdly, we discuss why $\text{COMA}^\text{scd}$ outperforms LRA. For details on the performance metrics, we refer to Appendix \ref{app:perfmetrics}.

\subsection{Performance of $\text{COMA}^\text{scd}$} \label{sec:results}

\begin{figure}[]
    \vspace{-4 mm}
    \includegraphics[width=\textwidth, keepaspectratio]{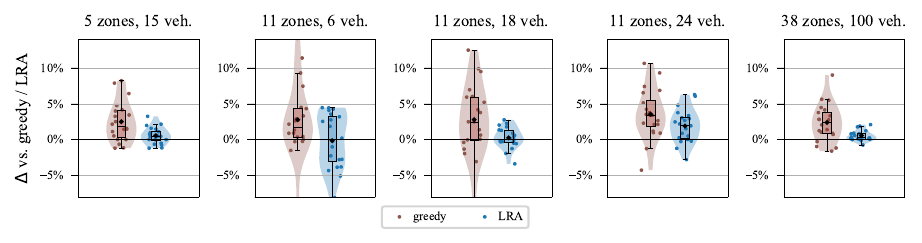}
    \vspace{-10 mm}
    \caption{Relative test performance $\Delta\ [\%]$ of $\text{COMA}^\text{scd}$ vs. greedy and LRA for multiple test dates.}
    \label{fig:plot_1}
    \vspace{-4 mm}
\end{figure}

\begin{table}[]
\setlength{\tabcolsep}{0.1cm}
    \fontsize{9pt}{9pt}\selectfont
    \centering
    \begin{tabular}{r|r|r|r|r|r}
        & 5 zones, 15 veh. & 11 zones, 6 veh. & 11 zones, 18 veh. & 11 zones, 24 veh. & 38 zones, 100 veh. \\ \hline
        vs. LRA & 0.5\% & -0.2\% & 0.2\% & 1.9\% & 0.6\% \\
        p-value & 0.05 & 0.52 & 0.22 & 0.00 & 0.00 \\
        vs. greedy & 2.5\% & 2.8\% & 2.8\% & 3.5\% & 2.4\% \\
        p-value & 0.00 & 0.00 & 0.01 & 0.00 & 0.00 \\
    \end{tabular}
    \vspace{-4 mm}
    \caption{Mean test performance improvement of $\text{COMA}^\text{scd}$ vs. LRA and greedy, including the respective Wilcoxon p-values.}
    \label{tab:main_table}
    \vspace{-8 mm}
\end{table}

We present results of our tests of $\text{COMA}^\text{scd}$ in Figure \ref{fig:plot_1} and Table \ref{tab:main_table}. In all instances except the one with 11 zones and 6 vehicles, $\text{COMA}^\text{scd}$ outperforms LRA and the greedy algorithm on average, the former by up to 1.9\% and the latter by up to 3.5\%. On single test dates, $\text{COMA}^\text{scd}$ can outperform LRA by up to 6\%. This improvement is significant, as the Wilcoxon p-values are at most 5\% for the respective instances. While these relative improvements appear to be small, they are substantial in AMoD and of a similar magnitude as the improvements of previous state-of-the-art algorithms over their respective benchmarks \citep[cf.][]{SadeghiEshkevari2022,Enders2023}.

In the instance with a high acceptance rate (5 zones, 15 vehicles), the significant performance improvement of $\text{COMA}^\text{scd}$ compared to LRA is an especially positive result, as a vehicle is usually available for each request in this instance, limiting the improvement potential for DRL. In the instance with a low acceptance rate (11 zones, 6 vehicles), the improvement potential is similarly limited, as vehicles are rarely idle. Consequently, the performance of $\text{COMA}^\text{scd}$ is most similar to LRA in this instance. In contrast, the instance of 11 zones and 24 vehicles has a balanced ratio between the number of vehicles and requests. Here, the performance improvement of $\text{COMA}^\text{scd}$ is the largest of all instances, outperforming LRA by on average 1.9\% and greedy by on average 3.5\%. In the large instance (38 zones, 100 vehicles), $\text{COMA}^\text{scd}$ significantly improves performance by on average 0.6\% compared to LRA, proving that the algorithm is applicable to large-scale environments. The lower performance improvement can be explained by the weight of $\text{COMA}^\text{adj}$ being required to increase more slowly in the loss function of $\text{COMA}^\text{scd}$ when the number of agents increases.

\subsection{Ablation study} \label{sec:ablation}
\begin{figure}[]
    \vspace{-8 mm}
    \includegraphics[width=\textwidth, keepaspectratio]{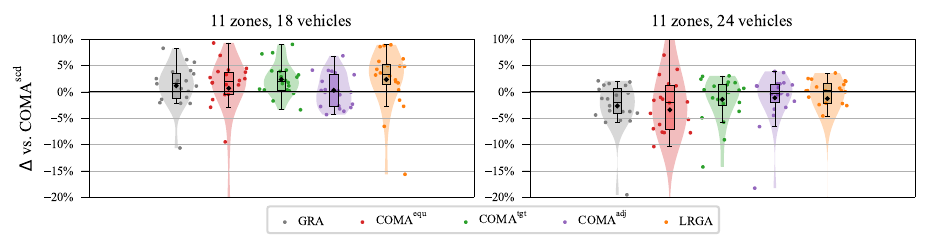}
    \vspace{-10 mm}
    \caption{Relative test performance $\Delta\ [\%]$ of all algorithms vs. $\text{COMA}^\text{scd}$.}
    \label{fig:plot_2}
    \vspace{1 mm}
\end{figure}

\begin{table}[]
\setlength{\tabcolsep}{0.1cm}
    \begin{minipage}[t]{0.57\textwidth}
    \fontsize{9pt}{9pt}\selectfont
    \centering
    \begin{tabular}[t]{r|c|c|c|c|c}
        & 5 zones & 11 zones & 11 zones & 11 zones & 38 zones \\
        & 15 veh. & 6 veh. & 18 veh. & 24 veh. & 100 veh. \\ \hline
        GRA & \checkmark & \checkmark & $\ovoid$ & $\ovoid$ & -- \\
        $\text{COMA}^\text{equ}$ & \checkmark & \checkmark & $\ovoid$ & $\ovoid$ & -- \\
        $\text{COMA}^\text{tgt}$ & \checkmark & \checkmark & $\ovoid$ & $\ovoid$ & -- \\
        $\text{COMA}^\text{adj}$ & \checkmark & \checkmark & $\ovoid$ & $\ovoid$ & -- \\
        LGRA & \checkmark & \checkmark & \checkmark & \checkmark & -- \\
        LRA & \checkmark & \checkmark & \checkmark & \checkmark & \checkmark \\
        $\text{COMA}^\text{scd}$ & \checkmark & \checkmark & \checkmark & \checkmark & \checkmark \\
    \end{tabular}
    \vspace{-3 mm}
    \caption{Convergence of algorithms. \checkmark\: denotes stable convergence, $\ovoid$ unstable convergence (across random seeds),\: --\: no convergence.}
    \label{tab:convergence}
    \end{minipage}
    \hfill
    \begin{minipage}[t]{0.39\textwidth}
    \fontsize{9pt}{9pt}\selectfont
    \centering
    \begin{tabular}[t]{r|r|r}
        & \multicolumn{2}{c}{11 zones} \\
        & 18 vehicles & 24 vehicles \\ \hline
        GRA & 1.2\% (0.02) & -2.7\% (0.01) \\
        $\text{COMA}^\text{equ}$ & 0.7\% (0.03) & -3.5\% (0.04) \\
        $\text{COMA}^\text{tgt}$ & 2.3\% (0.00) & -1.5\% (0.23) \\
        $\text{COMA}^\text{adj}$ & 0.3\% (0.42) & -1.2\% (0.41) \\
        LGRA & 2.3\% (0.00) & -1.3\% (0.55) \\
        \multicolumn{3}{c}{} \\
        \multicolumn{3}{c}{} \\
    \end{tabular}
    \vspace{-2.5 mm}
    \caption{Test performance of algorithms vs. $\text{COMA}^\text{scd}$ (Wilcoxon p-value).}
    \label{tab:ablation_table}
    \end{minipage}
    \hfill
    \vspace{-8 mm}
\end{table}

In the following, we discuss the performance of all proposed algorithms with respect to numerical stability and scalability across random seeds (Table~\ref{tab:convergence}) as well as computational performance (Table~\ref{tab:ablation_table} \& Figure~\ref{fig:plot_2}). As can be seen in Table~\ref{tab:convergence}, all algorithms show stable convergence for the small instances, while all but LRA, LGRA, and $\text{COMA}^\text{scd}$ exhibit stability issues already for medium-sized instances, failing to converge for one-third of the seeds and requiring about ten times as many training steps to converge compared to LRA for the remaining seeds. In contrast, LGRA and $\text{COMA}^\text{scd}$ converge on all seeds and require comparable to at maximum twice as many training steps compared to LRA. For the large instance, only $\text{COMA}^\text{scd}$ and the LRA baseline converge, with $\text{COMA}^\text{scd}$ requiring similar to twice as many training steps. Figure~\ref{fig:plot_2} and Table~\ref{tab:ablation_table} show the relative performance of all algorithms compared to $\text{COMA}^\text{scd}$ for the medium-sized instances over seeds for which all algorithms converged. As can be seen, results are mixed: while pure global-rewards-based algorithms outperform $\text{COMA}^\text{scd}$ on average on the 18 vehicles instance, $\text{COMA}^\text{scd}$ outperforms all other algorithms on the 24 vehicles instance. To understand this ambiguous effect, we detail the algorithms' convergence behavior: increasing the instance from 18 to 24 vehicles technically requires to train 570 instead of 430 agents, which significantly challenges all purely global-rewards-based algorithms. In fact, looking at the validation reward variance of each algorithm (see Appendix~\ref{app:training}), we observe converging but less stable learning behavior for all purely global-rewards-based algorithms, which explains the respective performance drop. While this observation manifests the robustness of $\text{COMA}^\text{scd}$ at a performance that improves upon local-rewards-based algorithms, it also points at a promising direction for future research: if one manages to stabilize and scale the purely global-rewards-based algorithms, one will most likely obtain even better performance.

\subsection{Structural analysis} \label{sec:analysis}
Finally, we aim to understand the performance difference between the studied algorithms by analyzing the respective policy characteristics in Table~\ref{tab:perf_inve}, which details request rejection rates of profitable requests for our LRA baseline, the pure global-rewards-based algorithm $\text{COMA}^\text{adj}$, and $\text{COMA}^\text{scd}$. Both COMA-based algorithms have a lower rejection rate compared to LRA, which explains their improved performance. This finding, as well as the relation between $\text{COMA}^\text{scd}$ and $\text{COMA}^\text{adj}$, are in line with the performance shown for the 24 vehicles instance in Figure~\ref{fig:plot_2}. To understand operational intricacies, we analyze the difference between average rejection rates of empty destination zones and zones that contain more than two vehicles upon a request's arrival, which is 3.8\% for LRA, 6.4\% for $\text{COMA}^\text{adj}$ and 4.9\% for $\text{COMA}^\text{scd}$. This indicates a stronger focus of $\text{COMA}^\text{scd}$ and $\text{COMA}^\text{adj}$ on implicit vehicle balancing, as these algorithms are more reluctant to send vehicles to already crowded zones. Such a focus on vehicle balancing stems from global reward structures and partially explains performance improvements: if vehicles are unbalanced, less overall requests can be served; individual vehicles might still obtain high local rewards, but the global reward decreases.

Beyond implicit balancing, we analyze the algorithms' anticipative performance, i.e., their capability to foresee future demand and consider it during decision-making. To do so, we analyze an algorithm's overperformance ratio (see Table~\ref{tab:perf_inve}), which compares the summed theoretical profits of future requests in the same zone following acceptance or rejection of an initially profitable request. We calculate this ratio by dividing the total theoretical profit after rejections by that after acceptances (see Appendix~\ref{app:overperformance} for details). As can be seen, $\text{COMA}^\text{adj}$ has the highest overperformance ratio, followed by LRA and $\text{COMA}^\text{scd}$. From the higher ratio, we conclude that $\text{COMA}^\text{adj}$ has better forecasting abilities, as requests after rejections are more profitable than requests after acceptances under its policy. In contrast, $\text{COMA}^\text{scd}$ appears to suffer from the mixture of local and global rewards, which can explain some of the performance gaps compared to algorithms with purely global rewards. A possible reason for the better forecasting abilities of $\text{COMA}^\text{adj}$ is that global Q-values incorporate more implicit information about future demand and thus more prescriptive information, making them more representative: since global rewards are less dependent on the actions of individual agents, it is easier to infer information about demand from them.

\begin{table}[]
    \vspace{-8 mm}
    \fontsize{9pt}{9pt}\selectfont
    \centering
    \begin{tabular}{l|r|r|r}
        measure & LRA & $\text{COMA}^\text{adj}$ & $\text{COMA}^\text{scd}$ \\
        \hline
        rejection rate (of profitable requests) & 17.6$\%$ & 16.6$\%$ & 15.5$\%$ \\
        \hspace{4mm} $\to$ rejection rate for destination zones without vehicles & 16.4$\%$ & 12.9$\%$ & 13.6$\%$ \\
        \hspace{4mm} $\to$ rejection rate for destination zones with $>$2 vehicles & 20.2$\%$ & 19.3$\%$ & 18.5$\%$ \\
        overperformance ratio (rejections / acceptances) & 1.75 & 1.87 & 1.27 \\
    \end{tabular}
    \vspace{-2 mm}
    \caption{Rejection rates of generally profitable requests on the instance with 24 vehicles.}
    \label{tab:perf_inve}
    \vspace{-4 mm}
\end{table}
\vspace{-0.05cm}

\section{Conclusion}
We study vehicle dispatching for AMoD systems, where a profit-maximizing central operator assigns vehicles to requests or rejects these. We propose a novel MADRL algorithm with global rewards and credit assignment based on a counterfactual baseline. Our algorithm combines the benefits of low learning variance and sample efficiency of SAC with the benefits of credit assignment of COMA. We show that a naive combination of SAC and COMA does not converge, and develop a stable and scalable algorithm that uses reward scheduling with a new advantage function based on an equally-weighted baseline and a target actor network baseline. We show that our algorithm improves upon the current state-of-the-art by up to 2\% on average across test dates and up to 6\% on individual dates from real-world data sets. This constitutes a substantial performance improvement for the AMoD application case where a single percent improvement yields significant daily monetary gains. We further provide a structural analysis which shows that the use of global rewards can improve implicit vehicle balancing and demand forecasting abilities. Our proposed algorithm is applicable beyond the area of AMoD, as it can be useful in any application where stable multi-agent deep reinforcement learning with credit assignment is required. An interesting avenue for future research would be the comparison of our algorithm to decentralized learning algorithms with local actors and critics. Future work may furthermore apply our algorithm to other areas or focus on stabilizing purely global-rewards-based algorithms.

\appendix

\section{Formal definition of problem setting}\label{app:problem}

In the following, we lay out the Markov Decision Process formally defining our problem setting. This Markov Decision Process, as our problem setting, is the same as the one used by \citet{Enders2023} to ensure comparability.

\paragraph{Preliminaries:} Our time horizon $\mathcal{T} = \{0,1,\dots,T\}$ is discrete. At every time step, customers can submit multiple requests, while the operator makes one decision per time step, if required deciding on multiple requests at once. The operating area is in our case represented by a graph $G=(V,E)$ with weight vectors $^e\bm{w}=(^ew^1,^ew^2) \in \mathbb{R}_{\>0} \times \mathbb{N}$, in which $^ew^1$ is the spatial length of an edge $e \in E$ and $^ew^2$ is the amount of time steps necessary to traverse it. The nodes of $G$ may represent the centers of service zones or dedicated locations depending on the respective case study data.

\paragraph{States:} The state of the system at time $t \in \mathcal{T}$ is described by
\begin{equation*}
    \bm{S}_t = \left( t,\: \left(^t\bm{r}^i\right)_{i \in \{1,\dots,R_t\}}, \:\left(\bm{k}^j_t\right)_{j \in \{1,\dots,K\}} \right),
\end{equation*}
where $R_t$ is the variable number of new requests $^t\bm{r}^i, i \in \{1,\dots,R_t\}$ at time step $t$. $K$ is the constant number of vehicles $\bm{k}^j_t, t \in \{1,\dots,K\}$. Requests $\bm{r} = (\omega, o, d)$ consist of a waiting time $\omega \in \mathbb{N}_0 \cup \emptyset$, an origin $o \in V$ and a destination $d \in V \backslash \{o\}$. The waiting time $\omega$ tracks how much time has elapsed between the submission of a request and the pickup of a customer, at which point we set $\omega \leftarrow \emptyset$. Vehicles $\bm{k}=(v,\tau,\bm{r}^1,\bm{r}^2)$ are described by a position $v \in V$, the number of steps $\tau \in \mathbb{N}_0$ to reach that position, as well as two requests $\bm{r}^1$ and $\bm{r}^2$. The position $v$ is either the current actual position if the vehicle has no requests assigned, or it is the node reached after serving all currently assigned requests. Each vehicle can be assigned to a maximum of two requests, more requests would be impractical for realistic trip lengths and waiting times. As we do not pool requests, vehicles serve their assigned requests sequentially. The position of vehicle $\bm{k}^j_t$ is denoted as $^jv_t$, the same holds for the other components of the vehicle vector.

\paragraph{Actions:} At each time step $t$, the central operator has the action space
\begin{equation}\label{eqn:action_space}
    \begin{split}
        \mathcal{A}(\bm{S}_t)=\Biggl\{\left(a^1_t,\dots,a^{R_t}_t\right)\Bigg|\:&a^i_t=0 \vee \left(a^i_t=j \in \{1,\dots,K\} \wedge\: ^j\bm{r}^2_t = \emptyset \right) \forall i \in \{1,\dots,R_t\}, \Biggr. \\
        &\left. \sum_{i=1}^{R_t} \mathbb{1}\left(a^i_t=j \right)\leq 1\forall j \in \{1,\dots,K\} \right\}.
    \end{split}
\end{equation}
For each request $^t\bm{r}^i, i \in \{1,\dots,R_t\}$, the operator can take one decision $a^i_t$, which is either its rejection $(a^i_t=0)$ or its assignment to one vehicle $\bm{k}^j(a^i_t=j)$. Assigning a request to a vehicle is only possible if a vehicle has less than two already assigned requests, formally if $^j\bm{r}^2_t=\emptyset$. If the operator rejects a request, the request leaves the system and cannot be accessed later on. As a realistic simplification of the matching algorithm, only one request can be assigned to each vehicle at one time step, as displayed in the last condition in Equation~\eqref{eqn:action_space}. At every time step, the action space of the central operator has a size of $(K+1)^{R_t}$.

\paragraph{Transitions:} The transition from one state to the next consists of two parts: At first, the transition from the pre-decision to the post-decision state, dependent on the action taken by the operator. Subsequently, the transition from the post-decision to the next pre-decision state, independent of the selected action.

Considering the first part of a transition, rejecting a request does not change the state. When accepting a request by assigning it to a vehicle $a^i=j$, the request is added to the state of the respective vehicle $j$. Specifically, $^j\bm{r}^1 \leftarrow{} ^t\bm{r}^i$ if $^j\bm{r}^1=\emptyset$, else $^j\bm{r}^2 \leftarrow{} ^t\bm{r}^i$. Within the second part of a transition, vehicles pick up customers, drop off customers and move between nodes to serve requests. When a vehicle picks up a customer, formally when $\bm{r}^1 \neq\emptyset\wedge\tau=0\wedge v=o(\bm{r}^1)$, $o(\bm{r}^1)$ denoting the origin of request $\bm{r}^1$, then $\omega(\bm{r}^1) \leftarrow \emptyset$. A movement between two nodes, taking place if $\tau>0$, is denoted as $\tau\leftarrow\tau-1$. Once a vehicle reaches its destination, but has at least one unfulfilled request, denoted as $\tau=0\wedge\bm{r}^1\neq\emptyset$, then the next node on the vehicle's path to serve the current request $v'$ replaces $v$. Additionally, $\tau\leftarrow{}^{(v,v')}w^2-1$. We shift requests, formally $\bm{r}^1\leftarrow\bm{r}^2$ and $\bm{r}^2\leftarrow\emptyset$, if a vehicle fulfills a request before the next decision, formally if $\bm{r}^1\neq\emptyset\wedge\omega(\bm{r}^1)=\emptyset \wedge\tau=0\wedge v=d(\bm{r}^1)$. We increase the waiting times of requests that have not been picked up so far, i.e., of requests that have $\omega\neq\emptyset$, by one: $\omega\leftarrow\omega+1$, $\omega$ referring to $\omega(\bm{r}^1)$ and/or $\omega(\bm{r}^2)$. Since customers, independent of the vehicles' states, place new requests, $(^{t+1}\bm{r}^i)_{i\in\{1,\dots,R_{t+1}\}}$ replaces $(^t\bm{r}^i)_{i\in\{1,\dots,R_t\}}$. The operator does not have knowledge of the probability distribution generating the requests over time, but it is possible to simulate this distribution using data of past requests. As a last step of each transition, $t\leftarrow t+1$.

\paragraph{Rewards:} Our reward function is the central operator's operating profit, since fixed costs do not depend on the dispatching problem. The operating profit is defined as the revenue from fulfilling requests, reduced by operational costs, e.g., for fuel or maintenance. Vehicles only incur costs when moving, as the lack of drivers' salaries makes the operating costs of idle vehicles negligible. Revenues are earned if a vehicle picks up a customer within the maximum waiting time. The operator has a pricing model, represented as a revenue function $\text{rev}(\bm{r})\in\mathbb{R}_{>0}$. To ease understanding, we denote the profit components dependent of the post-decision state $S_{t^+}$, writing $t$ for $t^+$. At time $t$, the total revenue then is defined as
\begin{equation*}
    \text{Rev}(\bm{S}_t)=\sum_{j=1}^K\mathds{1}\left(^j\bm{r}^1_t \neq\emptyset\wedge{}^j\tau_t=0\wedge{}^jv_t= o\left(^j\bm{r}^1_t\right) \wedge\omega\left(^j\bm{r}^1_t \right) \leq\omega^{max}\right)\cdot\text{rev}\left(^j\bm{r}^1_t\right).
\end{equation*}
Vehicles incur operational costs $c\in\mathbb{R}_{>0}$ per distance unit when starting to move from $v$ to $v'$, in line with common assumptions \citep[cf.][]{Boesch2018}. The total cost at time $t$ is thus defined as
\begin{equation*}
    \text{Cost}(\bm{S}_t)=c\cdot\sum_{j=1}^K\mathds{1}\left(^j\tau_t=0 \wedge{}^j\bm{r}^1_t \neq\emptyset\right)\cdot{} ^{(^jv_t,^jv'_t)}w^1.
\end{equation*}
The profit at time $t^+$ is calculated as $\text{Profit}(\bm{S}_{t^+})=\text{Rev}(\bm{S}_{t^+})-\text{Cost}(\bm{S}_{t^+})$. Since $\bm{S}_{t^+}$ is a function of the pre-decision state $\bm{S}_t$ and the selected action $\bm{a_t}\in\mathcal{A}(\bm{S}_t)$, we can write $\text{Profit}(\bm{S}_{t^+})=\text{Profit}(\bm{S}_t,\bm{a}_t)$. The goal of the system operator is to find a policy $\pi(\bm{a_t}|\bm{S}_t)$ which, given the initial state $\bm{S}_0$, maximizes the expected total reward over all time steps
\begin{equation*}
    \text{Profit*}(\bm{S}_0)=\max_\pi\mathbb{E}\left[\:\sum_{t=0}^{T-1}\text{Profit}(\bm{S}_t,\bm{a_t})\:\middle|\:\bm{S}_0\:\right].
\end{equation*}

\section{Proof of Proposition \ref{prop:equivalence}}\label{sec:proof}

The combined loss function of SAC and COMA for one instance of a batch and one agent is
\begin{equation*}
    J_\pi(\phi|s,i)=\sum_{a_i}\pi(a_i) \left(\alpha\log\pi(a_i)-Q(a_i)+\sum_{a'_i}\pi(a'_i)\: Q(a'_i) \right).
\end{equation*}
Considering the last summand separately, we get
\begin{align*}
    &\sum_{a_i}\pi(a_i)\sum_{a'_i}\pi(a'_i)\: Q(a'_i) \\
    &=\pi_1\:(\pi_1\: Q_1+\pi_2\: Q_2+...+\pi_n\: Q_n)
    +...+\pi_n\:(\pi_1\: Q_1+\pi_2\: Q_2+...+\pi_n\: Q_n) \\&=\pi^2_1\:Q_1+\pi_1\:\pi_2\:Q_2+...+\pi_1\:\pi_n\:Q_n+\pi_2\:\pi_1\:Q_1+...+\pi_n\:\pi_1\:Q_1+...+\pi^2_n\:Q_n \\
    &=\pi_1\:Q_1\:(\pi_1+\pi_2+...+\pi_n)+...+\pi_n\:Q_n\:(\pi_1+\pi_2+...+\pi_n) \\
    &=\sum_{a_i}\pi(a_i)\: Q(a_i).
\end{align*}
Inserting that into the loss function, we obtain
\begin{equation*}
        J_\pi(\phi|s,i)=\sum_{a_i}\pi(a_i)\: \alpha\log\pi(a_i)-\sum_{a_i}\pi(a_i)\: Q(a_i)+\sum_{a_i}\pi(a_i)\: Q(a_i)=\sum_{a_i}\pi(a_i)\: \alpha\log\pi(a_i),
\end{equation*}
which is just the entropy. If the loss function equals the entropy, the actor is not trained to maximize the probabilities of actions associated with the highest Q-values and thus cannot learn a meaningful policy. Note that this problem occurs for an arbitrary number of $n$ actions per agent (as displayed), including our setting with $n=2$ actions.

\section{Experiments}

\subsection{Experimental setup} \label{app:setup}

\begin{figure}[]
    \vspace{-6 mm}
    \includegraphics[width=\textwidth, keepaspectratio]{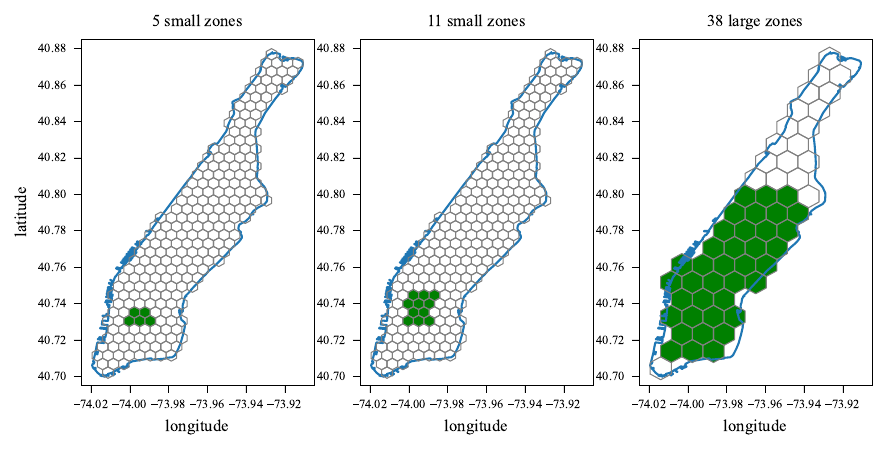}
    \vspace{-10 mm}
    \caption{Hexagonal grid of Manhattan with considered areas for instances of this paper marked in green.}
    \label{fig:zones_plot}
    \vspace{-4 mm}
\end{figure}

\paragraph{Data:} We use the same experimental setup as \citet{Enders2023}: our data consists of yellow taxi trip records from 2015, excluding weekends and holidays. We define the time of request placement as the pickup time in the data set, and consider a one hour time interval per day as one episode, using a time step size of one minute. We discretize the island of Manhattan using small or large hexagonal zones, as displayed in Figure \ref{fig:zones_plot}. Neighboring zones' centers have a distance of 459 or 917 meters for the small and large zones, respectively. We use discrete zones of the indicated sizes, because they allow us to test our algorithm within reasonable training times for small/medium and larger instances. The zones' hexagonal shape allows approximately realistic estimations of travel distances and times. In our instances, 5 small zones cover close to one square kilometer in central Manhattan, 11 small zones cover about two square kilometers including the area of the 5 small zones, and 38 large zones roughly cover the southern two thirds of Manhattan. The instances are depicted in Figure \ref{fig:zones_plot}. These instances allow us to test algorithms on small instances, e.g., for testing edge cases of the ratio between supply and demand (5 small zones), medium instances which are comparable to instances used in related work (11 small zones), and large instances which enable testing the scalability of algorithms (38 large zones). For each instance, we delete all requests originating in or leading to zones outside the area under consideration and all requests that start and end within the same zone. Vehicles travel along the shortest route between zones. We assume a travel time of two time steps between neighboring small zones and five time steps between neighboring large zones. The maximum waiting time of customers between request submission and pickup is five minutes. We set the revenue to 5.00~USD per km and the costs to 4.50~USD per km, resulting in a profit margin of 10\% in ideal cases. The maximum number of requests per time step never exceeds 7 for the instances with 5 small zones and 24 for the instances of 11 small zones. For the instances of 38 large zones, we use only every  $\text{20}^{\text{th}}$ request to make the problem size tractable for our available hardware. This results in a maximum of 20 requests per time step in these instances.

\paragraph{Hardware:} We train the models on NVIDIA Tesla P100, V100, or A100 GPUs. Depending on the problem size and number of time steps for training, this training takes between 2 and 66 hours.

\paragraph{Benchmark:} The greedy algorithm considers all requests in the order of their arrival. If a vehicle is available to serve the request with a positive profit and within the maximum waiting time, it is assigned to serve the request. If there are multiple available vehicles, the greedy algorithm assigns the vehicle that can serve the request with the highest profit. If no vehicle is available, the greedy algorithm rejects the request. In the instance with 5 zones and 15 vehicles, the greedy algorithm accepts 78\% of requests. This acceptance rate is 22\% in the instance with 11 zones and 6 vehicles, 47\% in the instance with 11 zones and 18 vehicles, 55\% in the instance with 11 zones and 24 vehicles, and 50\% in the instance with 38 zones and 100 vehicles.

\paragraph{Practical application:} When applying our algorithm in practice, we propose training the algorithm in a simulation and fine-tuning it using real-world experience. The simulation should reflect the geographical properties of the intended application area. For early training, simulated data based on real-world observations can be used as training and validation data. For later training, data collected in the intended application area should be used as training and testing data to ensure optimal performance.

\subsection{Hyperparameters} \label{app:hyperparameters}

We mostly use the same hyperparameters as \citet{Enders2023} for our algorithm, only adjusting the entropy coefficient $\alpha$, the learning rate, and the total number of training steps. A different number of random steps in the beginning, size of the replay buffer, batch size, or discount factor does not improve our results.

We train models until their validation performance is stable and does not increase further. Depending on the instance and algorithm, this results in 200,000 to 2,000,000 training steps. The number of required steps increases with the instance size, i.e., with the number of zones and vehicles. Across instances, LRA, $\text{COMA}^\text{scd}$ and LGRA require 200,000 to 400,000 steps, while GRA, $\text{COMA}^\text{equ}$, $\text{COMA}^\text{tgt}$ and $\text{COMA}^\text{adj}$ require 200,000 to 2,000,000 steps. We separate the learning rates for actor and critic, as global rewards and the inclusion of a baseline can change the models' convergence properties. Tuning the learning rates for each instance, we set both learning rates to $3\cdot10^{-4}$ for LRA and to values between $3\cdot10^{-4}$ and $2.4\cdot10^{-3}$ for GRA. For the algorithms based on COMA ($\text{COMA}^\text{equ}$, $\text{COMA}^\text{tgt}$, $\text{COMA}^\text{adj}$, $\text{COMA}^\text{scd}$), we set the learning rates to values between $6\cdot10^{-4}$ and $1.8\cdot10^{-3}$, with the actor having higher learning rates than the critic. We further tune the entropy coefficient $\alpha$ for each instance, with values ranging between 0.37 and 1.5.

\subsection{Performance metrics} \label{app:perfmetrics}

At the end of each training run, we test the model with the best validation performance on the test data. For each time step, we store the operator's profit of that time step $\text{Profit}(\bm{S}_t,\bm{a_t})$ as a score. We calculate the score of a test date as the sum of all scores on that test date, $\text{Score}_d=\sum_{t\in d}\text{Profit}(\bm{S}_t,\bm{a_t})$, with $d$ being the test date. We repeat each training run with three different random seeds and average test scores $\text{Score}_d$ over these random seeds, $\text{Score}_d=\frac{1}{n_{\text{rs}}}\sum_{\text{rs}}\text{Score}_d(\text{rs})$, with $\text{rs}$ denoting the random seed and $n_{\text{rs}}$ denoting the number of random seeds. If a run does not converge to a reasonable performance, we exclude it from the average. This only applies to some of the algorithms in the ablation study in Section \ref{sec:ablation}.

To evaluate the algorithms, we subtract per-day scores of algorithms we intend to compare and investigate their differences $\text{Delta}_d(\text{alg}_1,\text{alg}_2)=\text{Score}_d(\text{alg}_1)-\text{Score}_d(\text{alg}_2)$, with $\text{alg}$ denoting the algorithm. In our figures, we display distributions of these per-day differences and analyze whether their mean, median and quartiles are above or below zero. This provides a visual indication of the magnitude and significance of performance differences between the considered algorithms. In the tables, we display the means of the per-day differences, $\text{Mean}(\text{alg}_1,\text{alg}_2)=\frac{1}{n_{d}}\sum_{d}\text{Delta}_d(\text{alg}_1,\text{alg}_2)$, with $n_{d}$ denoting the number of test dates. Furthermore, we display the p-values of Wilcoxon tests of these means. These measures provide a numerical indication of mean performance differences and their significance. Together, the visual and numerical measures allow us to assess the magnitude and statistical significance of performance differences between algorithms, with the mean score difference $\text{Mean}(\text{alg}_1,\text{alg}_2)$ being the most important metric for performance comparisons.

\subsection{Training schedules} \label{app:schedules}

The policy loss function of $\text{COMA}^\text{adj}$ is a dynamic weighted average of the loss functions of $\text{COMA}^\text{equ}$ and $\text{COMA}^\text{tgt}$, with $\beta$ being the weight of the loss function of $\text{COMA}^\text{tgt}$. Over the course of the training, $\beta$ always starts at zero, ends at one, and can increase linearly or following a power function. By default, we use a simple linear function. Since $\text{COMA}^\text{tgt}$ shows better experimental performance than $\text{COMA}^\text{equ}$ (see Section \ref{sec:ablation}), we also test a power function with an exponent of 0.5 to increase the share of $\text{COMA}^\text{tgt}$ quicker than when using a linear schedule. In the instance of 11 zones and 18 vehicles, the maximum average performance of a model with a power function is 4.6\% lower than the maximum average performance of a model with a linear function. In the instance of 11 zones and 24 vehicles, this number is 2.8\%. We conclude that a linearly increasing $\beta$ works best in general.

The policy loss function of $\text{COMA}^\text{scd}$ is a dynamic weighted average of the loss functions of LRA and $\text{COMA}^\text{adj}$, with $\kappa$ being the weight of the loss function of $\text{COMA}^\text{adj}$. Over the course of the training, $\kappa$ always starts at zero and ends at one. It can increase linearly, following a power function, or it can jump from zero to one at any specified point during training. Since our target is to use global rewards as soon as possible in training, we test power functions and jumps increasing the weight of $\text{COMA}^\text{adj}$ quicker than if following a simple linear schedule. We present results of these tests in Table \ref{tab:training_schedules}. We observe four patterns in these tests: firstly, power functions generally lead to the best performance, which is most likely the result of their smoother transition from local to global rewards in comparison to sudden jumps. Secondly, quickly increasing power functions with exponents between 0.01 and 0.5 work best in all instances. As a result of these quickly increasing functions, the share of $\text{COMA}^\text{adj}$ in the loss function rises faster in the best models than it would when following a linear schedule. This shows that a quick transition to global rewards enables utilizing their benefits for a larger part of the training. Thirdly, the best exponents vary between instances, i.e., a per-instance tuning of the schedule to increase $\kappa$ can lead to visible performance increases. Finally, the best performing exponent increases with the instance size. The resulting slower transition from local to global rewards in large instances is in line with our results concerning the scalability problems of global-rewards-based algorithms. We thus conclude that power functions with quickly increasing exponents generally work best, with large instances requiring larger exponents than small instances.

\begin{table}[]
\vspace{-4 mm}
\setlength{\tabcolsep}{0.1cm}
    \fontsize{9pt}{9pt}\selectfont
    \centering
    \begin{tabular}{r|r|r|r}
        instance & schedule type & exponent/jump & vs. best \\ \hline
        \multirow{4}{*}{5 zones, 15 veh.} & \textbf{power} & \textbf{0.01} & \textbf{0.0\%} \\
        & power & 0.05 & -1.5\% \\
        & power & 0.25 & -1.6\% \\
        & jump & 0.01 & -1.9\% \\ \hline
        \multirow{5}{*}{11 zones, 6 veh.} & \textbf{power} & \textbf{0.25} & \textbf{0.0\%} \\
        & jump & 0.01 & -0.7\% \\
        & power & 0.125 & -1.6\% \\
        & jump & 0.02 & -2.1\% \\
        & power & 0.01 & -4.4\% \\ \hline
        \multirow{7}{*}{11 zones, 18 veh.} & \textbf{power} & \textbf{0.25} & \textbf{0.0\%} \\
        & jump & 0.1 & -0.4\% \\
        & jump & 0.25 & -0.4\% \\
        & power & 2.00 & -0.6\% \\
        & power & 1.00 & -1.2\% \\
        & power & 0.50 & -2.1\% \\
        & jump & 0.50 & -3.1\% \\ \hline
        \multirow{4}{*}{11 zones, 24 veh.} & \textbf{power} & \textbf{0.25} & \textbf{0.0\%} \\
        & jump & 0.125 & -1.0\% \\
        & power & 0.125 & -1.6\% \\
        & power & 1.00 & -1.7\% \\ \hline
        \multirow{5}{*}{38 zones, 100 veh.} & \textbf{power} & \textbf{0.5} & \textbf{0.0\%} \\
        & power & 1.00 & -0.2\% \\
        & power & 0.25 & -0.3\% \\
        & jump & 0.25 & -0.3\% \\
    \end{tabular}
    \caption{Performance of training schemes of $\text{COMA}^\text{scd}$, relative to the best performance per instance. The specifications used in Sections \ref{sec:results} to \ref{sec:analysis} are displayed in bold writing. The column exponent/jump denotes the exponent in case of a power function or the jump point relative to the number of training steps in case of a jump schedule. For example, if the jump point is 0.25 and the model is trained for 400,000 steps, $\kappa$ is zero for the first 100,000 steps and one for the remainder of the training.}
    \label{tab:training_schedules}
\vspace{-6 mm}
\end{table}

\subsection{Validation rewards} \label{app:training}

\begin{figure}[]
    \includegraphics[width=\textwidth, keepaspectratio]{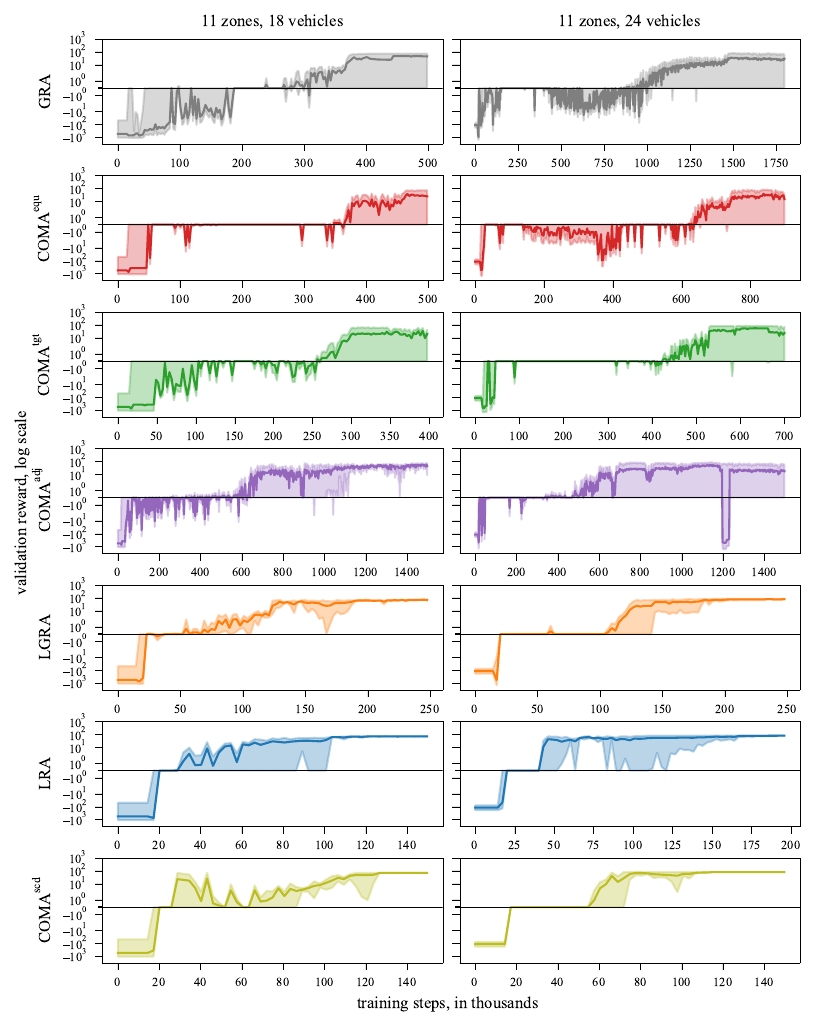}
    \vspace{-10 mm}
    \caption{Validation rewards of algorithms over training steps. The black horizontal line indicates zero. The main line denotes the average validation reward over the three random seeds, the shaded area the maximum and minimum rewards at each step.}
    \label{fig:convergence_plots}
\end{figure}

We show the validation rewards of GRA, $\text{COMA}^\text{equ}$, $\text{COMA}^\text{tgt}$, $\text{COMA}^\text{adj}$, LGRA, LRA, and $\text{COMA}^\text{scd}$ for the instance with 11 zones and 18 vehicles and the instance with 11 zones and 24 vehicles over the course of training in Figure \ref{fig:convergence_plots}. We observe three patterns in the validation rewards: firstly, in both instances, the purely global-rewards-based algorithms GRA, $\text{COMA}^\text{equ}$, $\text{COMA}^\text{tgt}$, and $\text{COMA}^\text{adj}$ need about ten times as many training steps to converge as LRA, while LGRA and $\text{COMA}^\text{scd}$ need about the same to double the number of steps until convergence. Secondly, most purely global-rewards-based algorithms display far larger differences between their best and worst validation performance in both instances, indicating unstable convergence behavior. Finally, the convergence speed decreases for the purely global-rewards-based algorithms except $\text{COMA}^\text{adj}$ in the instance with 24 vehicles compared to the one with 18 vehicles. For $\text{COMA}^\text{adj}$, the maximum and minimum performance diverges more in the larger instance. The algorithm LGRA also faces challenges when increasing the number of agents: to remain stable and converge quickly, we have to decrease the share of global rewards from 60\% to 30\%, thereby also lowering the positive influence of global rewards. In contrast, LRA and $\text{COMA}^\text{scd}$ have about the same stability and convergence speed in both instances.

This final observation provides evidence for the lower performance of purely global-rewards-based algorithms in the instance with 24 vehicles. As the learning is less stable, the learned policies are less reliable. With even validation performance curves of converging models being less stable, trained models are more likely to converge to a sub-optimal policy. While we allow all algorithms enough steps to converge, the slower learning of purely global-rewards-based algorithms can be an additional problem in practice. Overall, these results confirm our conclusion that learning using purely global rewards increases vehicle dispatching performance, but leads to problems when increasing the number of agents.

\subsection{Calculation of overperformance ratio} \label{app:overperformance}

We calculate the overperformance ratio as follows: firstly, we store the maximum profit of a request if it is positive together with the origin zone of this request. This is the profit that would be obtained if the request was served by the closest vehicle. Secondly, we store the theoretical profits that would be obtained from subsequent requests originating in the same zone if a vehicle was available at the same position as the vehicle closest to the original request. We store the theoretical profits for the ten time steps after the original request appears. Thirdly, we obtain the overprofits by subtracting the profit of the original request from the theoretical profits. We only count positive overprofits subsequently, as we only consider requests that would be more profitable than the original request. We sum these overprofits conditional on whether the original request is accepted or rejected. Finally, we divide the overprofit after a rejection by the overprofit after an acceptance.

A numerical illustration is as follows: consider a request, which could be served by a vehicle in the same zone with an initial profit of 10~USD, but the operator rejects this request. From the same zone, three further requests appear within the ten subsequent time steps. Using a vehicle within the same zone, they could be served with a profit of 5~USD, 12~USD, and 14~USD, respectively. These numbers are the theoretical profits. The associated overprofits are 0~USD, 2~USD, and 4~USD. Therefore, the total overprofit compared to the original request is 6~USD. Now, consider a second request, which is accepted by the operator, with an initial profit of 10~USD and subsequent theoretical profits of 8~USD, 11~USD, and 12~USD. The overperformance for this request is 3~USD. If these are the only two profitable requests available, the overperformance ratio is 6 divided by 3, which equals 2. As the overperformance ratio is larger than one, the subsequent requests after the rejection are more profitable than those after the acceptance. Accordingly, whenever the overperformance rato is larger than one, we can conclude that the analyzed algorithm is better at taking anticipative decisions than a greedy algorithm.

\acks{We thank the CORAIL research group at Polytechnique Montréal for valuable comments and discussions. The work of Heiko Hoppe was supported by a scholarship from the Max Weber-Programm Bayern.}

\bibliography{references.bib}

\begin{thebibliography}{35}
\providecommand{\natexlab}[1]{#1}
\providecommand{\url}[1]{\texttt{#1}}
\expandafter\ifx\csname urlstyle\endcsname\relax
  \providecommand{\doi}[1]{doi: #1}\else
  \providecommand{\doi}{doi: \begingroup \urlstyle{rm}\Url}\fi

\bibitem[Akkerman et~al.(2024)Akkerman, Luy, van Heeswijk, and
  Schiffer]{Akkerman2024}
Fabian Akkerman, Julius Luy, Wouter van Heeswijk, and Maximilian Schiffer.
\newblock {D}ynamic {N}eighborhood {C}onstruction for {S}tructured {L}arge
  {D}iscrete {A}ction {S}paces.
\newblock arXiv preprint at arXiv, 2024.
\newblock URL \url{arXiv:2305.19891}.

\bibitem[Alonso-Mora et~al.(2017)Alonso-Mora, Wallar, and Rus]{AlonsoMora2017}
Javier Alonso-Mora, Alex Wallar, and Daniela Rus.
\newblock Predictive routing for autonomous mobility-on-demand systems with
  ride-sharing.
\newblock In \emph{2017 {IEEE}/{RSJ} International Conference on Intelligent
  Robots and Systems ({IROS})}, 2017.

\bibitem[Bösch et~al.(2018)Bösch, Becker, Becker, and Axhausen]{Boesch2018}
Patrick~M. Bösch, Felix Becker, Henrik Becker, and Kay~W. Axhausen.
\newblock Cost-based analysis of autonomous mobility services.
\newblock \emph{Transport Policy}, 64, 2018.

\bibitem[Chang et~al.(2003)Chang, Ho, and Kaelbling]{Chang2003}
Yu-Han Chang, Tracey Ho, and Leslie Kaelbling.
\newblock All learning is {L}ocal: {M}ulti-agent {L}earning in {G}lobal
  {R}eward {G}ames.
\newblock In \emph{Advances in Neural Information Processing Systems
  ({N}eur{IPS})}, volume~16, 2003.

\bibitem[Christodoulou(2019)]{Christodoulou2019}
Petros Christodoulou.
\newblock Soft {A}ctor-{C}ritic for {D}iscrete {A}ction {S}ettings.
\newblock arXiv preprint at arXiv, 2019.
\newblock URL \url{arXiv:1910.07207v2}.

\bibitem[Enders et~al.(2023)Enders, Harrison, Pavone, and Schiffer]{Enders2023}
Tobias Enders, James Harrison, Marco Pavone, and Maximilian Schiffer.
\newblock Hybrid {M}ulti-agent {D}eep {R}einforcement {L}earning for
  {Au}tonomous {M}obility on {D}emand {S}ystems.
\newblock In \emph{Proceedings of The 5th Annual Learning for Dynamics and
  Control Conference ({L}4{DC})}, volume 211 of \emph{Proceedings of Machine
  Learning Research ({PMLR})}, 2023.

\bibitem[Foerster et~al.(2018)Foerster, Farquhar, Afouras, Nardelli, and
  Whiteson]{Foerster2018}
Jakob Foerster, Gregory Farquhar, Triantafyllos Afouras, Nantas Nardelli, and
  Shimon Whiteson.
\newblock {C}ounterfactual {M}ulti-{A}gent {P}olicy {G}radients.
\newblock \emph{Proceedings of the {AAAI} Conference on Artificial
  Intelligence}, 32\penalty0 (1), 2018.

\bibitem[Haarnoja et~al.(2018)Haarnoja, Zhou, Abbeel, and Levine]{Haarnoja2018}
Tuomas Haarnoja, Aurick Zhou, Pieter Abbeel, and Sergey Levine.
\newblock Soft {A}ctor-{C}ritic: {O}ff-{P}olicy {M}aximum {E}ntropy {D}eep
  {R}einforcement {L}earning with a {S}tochastic {A}ctor.
\newblock In \emph{Proceedings of the 35th International Conference on Machine
  Learning ({ICML})}, 2018.

\bibitem[Hadfield-Menell et~al.(2017)Hadfield-Menell, Milli, Abbeel, Russell,
  and Dragan]{HadfieldMenell2017}
Dylan Hadfield-Menell, Smitha Milli, Pieter Abbeel, Stuart~J. Russell, and Anca
  Dragan.
\newblock Inverse {R}eward {D}esign.
\newblock In \emph{Advances in Neural Information Processing Systems
  ({N}eur{IPS})}, volume~30, 2017.

\bibitem[He et~al.(2022)He, Wang, Han, Zou, and Miao]{He2022}
Sihong He, Yue Wang, Shuo Han, Shaofeng Zou, and Fei Miao.
\newblock A {R}obust and {C}onstrained {M}ulti-{A}gent {R}einforcement
  {L}earning {E}lectric {V}ehicle {R}ebalancing {M}ethod in {AM}o{D} {S}ystems.
\newblock arXiv preprint at arXiv, 2022.
\newblock URL \url{arXiv:2209.08230}.

\bibitem[Jiao et~al.(2021)Jiao, Tang, Qin, Li, Zhang, Zhu, and Ye]{Jiao2021}
Yan Jiao, Xiaocheng Tang, Zhiwei~(Tony) Qin, Shuaiji Li, Fan Zhang, Hongtu Zhu,
  and Jieping Ye.
\newblock Real-world ride-hailing vehicle repositioning using deep
  reinforcement learning.
\newblock \emph{Transportation Research Part C: Emerging Technologies}, 130,
  2021.

\bibitem[Jungel et~al.(2023)Jungel, Parmentier, Schiffer, and
  Vidal]{Jungel2023}
Kai Jungel, Axel Parmentier, Maximilian Schiffer, and Thibaut Vidal.
\newblock Learning-based {O}nline {O}ptimization for {A}utonomous
  {M}obility-on-{D}emand {F}leet {C}ontrol.
\newblock arXiv preprint at arXiv, 2023.
\newblock URL \url{arXiv:2302.03963}.

\bibitem[Kok and Vlassis(2006)]{Kok2006}
Jelle~R. Kok and Nikos Vlassis.
\newblock Collaborative {M}ultiagent {R}einforcement {L}earning by {P}ayoff
  {P}ropagation.
\newblock \emph{Journal of Machine Learning Research ({JMLR})}, 7, 2006.

\bibitem[Lee et~al.(2004)Lee, Wang, Cheu, and Teo]{Lee2004}
Der-Horng Lee, Hao Wang, Ruey~Long Cheu, and Siew~Hoon Teo.
\newblock Taxi {D}ispatch {S}ystem {B}ased on {C}urrent {D}emands and
  {R}eal-{T}ime {T}raffic {C}onditions.
\newblock \emph{Transportation Research Record: Journal of the Transportation
  Research Board}, 1882\penalty0 (1), 2004.

\bibitem[Liang et~al.(2022)Liang, Wen, Lam, Sumalee, and Zhong]{Liang2022}
Enming Liang, Kexin Wen, William H.~K. Lam, Agachai Sumalee, and Renxin Zhong.
\newblock An {I}ntegrated {R}einforcement {L}earning and {C}entralized
  {P}rogramming {A}pproach for {O}nline {T}axi {D}ispatching.
\newblock \emph{{IEEE} Transactions on Neural Networks and Learning Systems},
  33\penalty0 (9), 2022.

\bibitem[Liao(2003)]{Liao2003}
Ziqi Liao.
\newblock Real-time taxi dispatching using {G}lobal {P}ositioning {S}ystems.
\newblock \emph{Communications of the {ACM}}, 46\penalty0 (5), 2003.

\bibitem[Lin et~al.(2018)Lin, Beling, and Cogill]{Lin2018}
Xiaomin Lin, Peter~A. Beling, and Randy Cogill.
\newblock Multiagent {I}nverse {R}einforcement {L}earning for {T}wo-{P}erson
  {Z}ero-{S}um {G}ames.
\newblock \emph{IEEE Transactions on Games}, 10\penalty0 (1), 2018.

\bibitem[Meneses-Cime et~al.(2022)Meneses-Cime, Aksun~Guvenc, and
  Guvenc]{MenesesCime2022}
Karina Meneses-Cime, Bilin Aksun~Guvenc, and Levent Guvenc.
\newblock Optimization of {O}n-{D}emand {S}hared {A}utonomous {V}ehicle
  {D}eployments {U}tilizing {R}einforcement {L}earning.
\newblock \emph{Sensors}, 22\penalty0 (21), 2022.

\bibitem[Ng and Russell(2000)]{Ng2000}
Andrew~Y. Ng and Stuart~J. Russell.
\newblock Algorithms for {I}nverse {R}einforcement {L}earning.
\newblock In \emph{Proceedings of the Seventeenth International Conference on
  Machine Learning ({ICML})}, 2000.

\bibitem[Nguyen et~al.(2018)Nguyen, Kumar, and Lau]{Nguyen2018}
Duc~Thien Nguyen, Akshat Kumar, and Hoong~Chuin Lau.
\newblock Credit {A}ssignment {F}or {C}ollective {M}ultiagent {RL} {W}ith
  {G}lobal {R}ewards.
\newblock In \emph{Advances in Neural Information Processing Systems
  ({N}eur{IPS})}, volume~31, 2018.

\bibitem[{NYC TLC}(2015)]{NYCTLC2015}
{NYC TLC}.
\newblock Trip {R}ecord {D}ata.
\newblock Online, 2015.
\newblock URL \url{htps://www.nyc.gov/site/tlc/
  about/tlc-trip-record-data.page}.

\bibitem[Rashid et~al.(2020)Rashid, Samvelyan, De~Witt, Farquhar, Foerster, and
  Whiteson]{Rashid2020}
Tabish Rashid, Mikayel Samvelyan, Christian~Schroeder De~Witt, Gregory
  Farquhar, Jakob Foerster, and Shimon Whiteson.
\newblock Monotonic {V}alue {F}unction {F}actorisation for {D}eep
  {M}ulti-{A}gent {R}einforcement {L}earning.
\newblock \emph{Journal of Machine Learning Research ({JMLR})}, 21, 2020.

\bibitem[Sadeghi~Eshkevari et~al.(2022)Sadeghi~Eshkevari, Tang, Qin, Mei,
  Zhang, Meng, and Xu]{SadeghiEshkevari2022}
Soheil Sadeghi~Eshkevari, Xiaocheng Tang, Zhiwei Qin, Jinhan Mei, Cheng Zhang,
  Qianying Meng, and Jia Xu.
\newblock Reinforcement {L}earning in the {W}ild: {S}calable {RL} {D}ispatching
  {A}lgorithm {D}eployed in {R}idehailing {M}arketplace.
\newblock In \emph{Proceedings of the 28th {ACM} {SIGKDD} Conference on
  Knowledge Discovery and Data Mining}, 2022.

\bibitem[Son et~al.(2019)Son, Kim, Kang, Hostallero, and Yi]{Son2019}
Kyunghwan Son, Daewoo Kim, Wan~Ju Kang, David~Earl Hostallero, and Yung Yi.
\newblock {QTRAN}: {L}earning to {F}actorize with {T}ransformation for
  {C}ooperative {M}ulti-{A}gent {R}einforcement {L}earning.
\newblock In \emph{Proceedings of the 36th International Conference on Machine
  Learning ({ICML})}, 2019.

\bibitem[Su et~al.(2021)Su, Adams, and Beling]{Su2021}
Jianyu Su, Stephen Adams, and Peter Beling.
\newblock Value-{D}ecomposition {M}ulti-{A}gent {A}ctor-{C}ritics.
\newblock \emph{Proceedings of the {AAAI} Conference on Artificial
  Intelligence}, 35\penalty0 (13), 2021.

\bibitem[Sunehag et~al.(2018)Sunehag, Lever, Gruslys, Czarnecki, Zambaldi,
  Jaderberg, Lanctot, Sonnerat, Leibo, Tuyls, and Graepel]{Sunehag2018}
Peter Sunehag, Guy Lever, Audrunas Gruslys, Wojciech~Marian Czarnecki, Vinicius
  Zambaldi, Max Jaderberg, Marc Lanctot, Nicolas Sonnerat, Joel~Z. Leibo, Karl
  Tuyls, and Thore Graepel.
\newblock Value-{D}ecomposition {N}etworks {F}or {C}ooperative {M}ulti-{A}gent
  {L}earning {B}ased {O}n {T}eam {R}eward.
\newblock In \emph{Proceedings of the 17th International Conference on
  Autonomous Agents and MultiAgent Systems}, 2018.

\bibitem[Tang et~al.(2019)Tang, Qin, Zhang, Wang, Xu, Ma, Zhu, and
  Ye]{Tang2019}
Xiaocheng Tang, Zhiwei~(Tony) Qin, Fan Zhang, Zhaodong Wang, Zhe Xu, Yintai Ma,
  Hongtu Zhu, and Jieping Ye.
\newblock A {D}eep {V}alue-network {B}ased {A}pproach for {M}ulti-{D}river
  {O}rder {D}ispatching.
\newblock In \emph{Proceedings of the 25th {ACM} {SIGKDD} International
  Conference on Knowledge Discovery {\&} Data Mining}, 2019.

\bibitem[Tsao et~al.(2018)Tsao, Iglesias, and Pavone]{Tsao2018}
Matthew Tsao, Ramon Iglesias, and Marco Pavone.
\newblock Stochastic {M}odel {P}redictive {C}ontrol for {A}utonomous {M}obility
  on {D}emand.
\newblock In \emph{2018 21st International Conference on Intelligent
  Transportation Systems ({ITSC})}, 2018.

\bibitem[Wang et~al.(2018)Wang, Qin, Tang, Ye, and Zhu]{Wang2018}
Zhaodong Wang, Zhiwei Qin, Xiaocheng Tang, Jieping Ye, and Hongtu Zhu.
\newblock Deep {R}einforcement {L}earning with {K}nowledge {T}ransfer for
  {O}nline {R}ides {O}rder {D}ispatching.
\newblock In \emph{2018 {IEEE} International Conference on Data Mining
  ({ICDM})}, 2018.

\bibitem[Wei{\ss}(1995)]{Weiss1995}
Gerhard Wei{\ss}.
\newblock {D}istributed {R}einforcement {L}earning.
\newblock \emph{The Biology and Technology of Intelligent Autonomous Agents},
  1995.

\bibitem[Wolpert and Tumer(1999)]{Wolpert1999}
David~H. Wolpert and Kagan Tumer.
\newblock An {I}ntroduction to {C}ollective {I}ntelligence.
\newblock Techreport NASA-ARC-IC-99-63, NASA, 1999.

\bibitem[Wolpert and Tumer(2001)]{WolpertTumer2001}
David~H. Wolpert and Kagan Tumer.
\newblock {O}ptimal {P}ayoff {F}unctions {F}or {M}embers {O}f {C}ollectives.
\newblock \emph{Advances in Complex Systems}, 04\penalty0 (02n03), 2001.

\bibitem[Wu et~al.(2018)Wu, Rajeswaran, Duan, Kumar, Bayen, Kakade, Mordatch,
  and Abbeel]{Wu2018}
Cathy Wu, Aravind Rajeswaran, Yan Duan, Vikash Kumar, Alexandre~M. Bayen, Sham
  Kakade, Igor Mordatch, and Pieter Abbeel.
\newblock Variance {R}eduction for {P}olicy {G}radient with
  {A}ction-{D}ependent {F}actorized {B}aselines.
\newblock In \emph{6th International Conference on Learning Representations
  ({ICLR})}, 2018.

\bibitem[Xu et~al.(2018)Xu, Li, Guan, Zhang, Li, Nan, Liu, Bian, and
  Ye]{Xu2018}
Zhe Xu, Zhixin Li, Qingwen Guan, Dingshui Zhang, Qiang Li, Junxiao Nan,
  Chunyang Liu, Wei Bian, and Jieping Ye.
\newblock Large-{S}cale {O}rder {D}ispatch in {O}n-{D}emand {R}ide-{H}ailing
  {P}latforms.
\newblock In \emph{Proceedings of the 24th {ACM} {SIGKDD} International
  Conference on Knowledge Discovery {\&} Data Mining}, 2018.

\bibitem[Zhang et~al.(2017)Zhang, Hu, Min, Wu, Zhang, Feng, Gong, and
  Ye]{Zhang2017}
Lingyu Zhang, Tao Hu, Yue Min, Guobin Wu, Junying Zhang, Pengcheng Feng,
  Pinghua Gong, and Jieping Ye.
\newblock A {T}axi {O}rder {D}ispatch {M}odel based {O}n {C}ombinatorial
  {O}ptimization.
\newblock In \emph{Proceedings of the 23rd {ACM} {SIGKDD} International
  Conference on Knowledge Discovery and Data Mining}, 2017.

\end{thebibliography}

\end{document}